\documentclass[10pt,twocolumn,letterpaper]{article}

\usepackage[pagenumbers]{cvpr} 

\usepackage[accsupp]{axessibility}
\usepackage{graphicx}
\usepackage{amsmath}
\usepackage{amssymb}
\usepackage{booktabs}
\usepackage{multirow}
\usepackage{algorithm}
\usepackage{algorithmic}
\usepackage{listings}
\usepackage{stfloats}
\usepackage[pagebackref,breaklinks,colorlinks]{hyperref}

\usepackage[capitalize]{cleveref}
\crefname{section}{Sec.}{Secs.}
\Crefname{section}{Section}{Sections}
\Crefname{table}{Table}{Tables}
\crefname{table}{Tab.}{Tabs.}

\def\cvprPaperID{4734} 
\def\confName{CVPR}
\def\confYear{2022}

\begin{document}

\title{Brain-inspired Multilayer Perceptron with Spiking Neurons}

\author{Wenshuo Li$^{1}$, Hanting Chen$^{1,2}$, Jianyuan Guo$^{1}$, Ziyang Zhang$^{1}$, Yunhe Wang$^{1}$\thanks{Corresponding author}\\
$^{1}$Huawei Noah's Ark Lab.\\
$^{2}$Key Lab of Machine Perception (MOE), Dept. of Machine Intelligence, Peking University.\\
{\tt\small liwenshuo@huawei.com, yunhe.wang@huawei.com}
}
\maketitle

\begin{abstract}
Recently, Multilayer Perceptron (MLP) becomes the hotspot in the field of computer vision tasks. Without inductive bias, MLPs perform well on feature extraction and achieve amazing results. However, due to the simplicity of their structures, the performance highly depends on the local features communication machenism. To further improve the performance of MLP, we introduce information communication mechanisms from brain-inspired neural networks. Spiking Neural Network (SNN) is the most famous brain-inspired neural network, and achieve great success on dealing with sparse data. Leaky Integrate and Fire (LIF) neurons in SNNs are used to communicate between different time steps. In this paper, we incorporate the machanism of LIF neurons into the MLP models, to achieve better accuracy without extra FLOPs. We propose a full-precision LIF operation to communicate between patches, including horizontal LIF and vertical LIF in different directions. We also propose to use group LIF to extract better local features. With LIF modules, our SNN-MLP model achieves 81.9\%, 83.3\% and 83.5\% top-1 accuracy on ImageNet dataset with only 4.4G, 8.5G and 15.2G FLOPs, respectively, which are state-of-the-art results as far as we know. The source code will be available at \href{https://gitee.com/mindspore/models/tree/master/research/cv/snn_mlp}{$https://gitee.com/mindspore/models/tree/master/\\research/cv/snn\_mlp$}.
\end{abstract}

\section{Introduction}

\label{sec:intro}
\begin{figure*}[t]
\centering
\vspace{-12pt}
\includegraphics[width=5.6in]{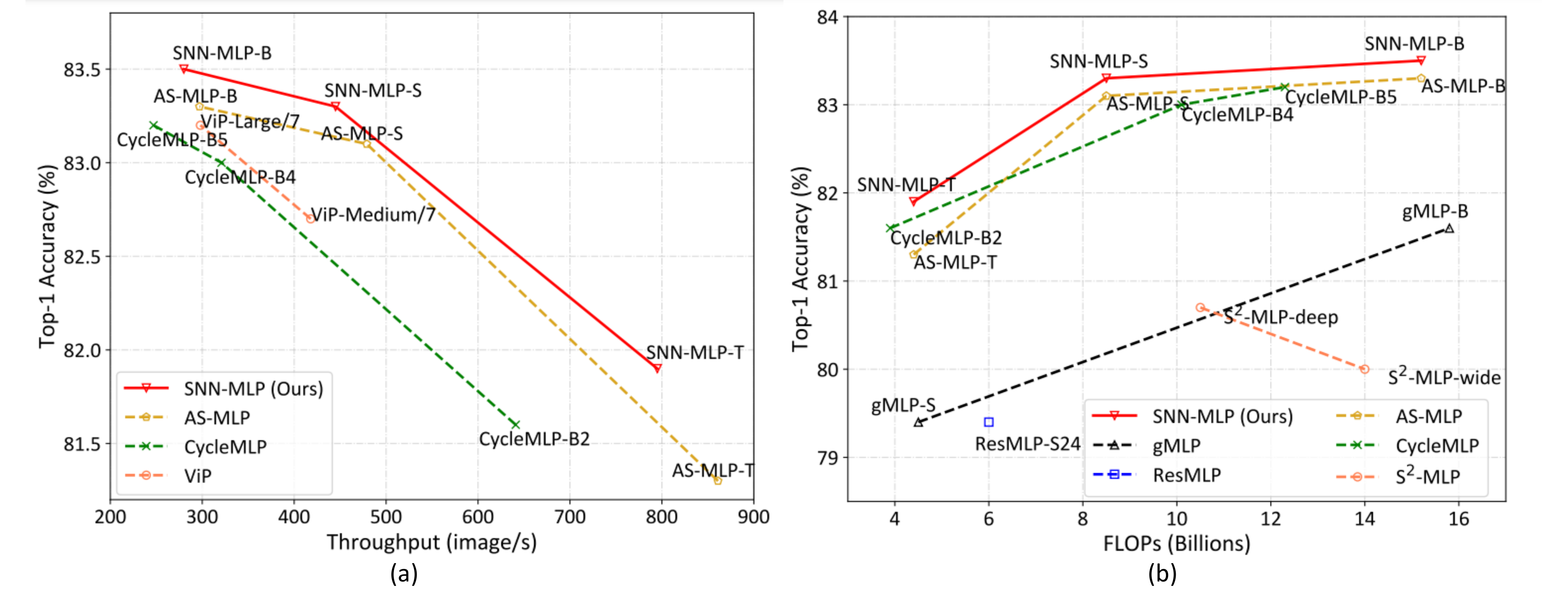}
\vspace{-10pt}
\caption{(a) FLOPs-Accuracy Pareto frontier. (b) Throughput-Accuracy Pareto frontier. The proposed SNN-MLP achieves the best results among these MLPs on both FLOPs-Accuracy and Throughput-Accuracy trade-off.}
\vspace{-15pt}
\label{fig:pareto-frontier}
\end{figure*}
With the help of inductive bias, convolution neural network (CNN) has become the most popular algorithm in several computer vision tasks, including image classification~\cite{he2016deep}, semantic segmentation~\cite{xiao2018unified} and object detection~\cite{he2017mask, chen2019hybrid}. CNN is easier to train and has fewer parameters compared with MLP, but inductive bias also limits its learning ability. Nowadays, CNNs are facing challenges from new types of vision backbones, like Transformers and MLPs. Transformers~\cite{vaswani2017attention} are initially proposed in the area of Nature Language Process (NLP), and researchers find that the self-attention machenism is also suitable for computer vision tasks~\cite{dosovitskiy2020image}. MLP receives wide-spread attention in academics, researchers find that easy operations like MLPs are as good as self-attention module. Without inductive bias, MLPs show better learning ability on larger datasets. The key point to utilize Transformers and MLPs is to divide images into patches and then apply the calculation to each patch. At present, one of the hotspots of research is how to communicate between patches. Permutation~\cite{hou2021vision} and shift operation~\cite{lian2021mlp, yu2021s} are the most common choice, and they all achieve admirable results.

In terms of information communication, SNN~\cite{maass1997networks} has a mature mechanism to deal with it. SNN is a kind of brain-inspired neural networks, and are frequently used to deal with sparse data, such as dynamic vision sensor (DVS)~\cite{kim2021optimizing}. The energy efficiency of SNN is highly competitive while SNN suffers from accuracy loss compared with CNN. The transformation from CNN/ANN to SNN often means up to 10\% accuracy drop on ImageNet and large time steps up to hundreds and thousands. Recently, the researches on SNN have been developed into two tracks. One shows how to transform CNN to SNN more efficiently and lossless~\cite{rueckauer2017conversion, sengupta2019going}. The other shows how to train SNN directly to achieve comparable accuracy with CNN~\cite{wu2019direct, zheng2020going}. Now state-of-the-art ANN-SNN conversion could adapt to classic CNN models with only 1\%-2\% accuracy drop while the time step is larger than 1000~\cite{sengupta2019going}. And state-of-the-art SNN training method could achieve 5\%-8\% accuracy drop with less than 10 time steps~\cite{wu2019direct, zheng2020going}. More time steps mean larger latency, so the performance of SNNs on general vision dataset like ImageNet is still not satisfying.

As we mentioned before, spiking neuron is used to communicate between different time step. In this paper, we introduce the brain-inspired spiking neurons (i.e. the LIF module in our paper) to communicate the information between patches in the MLP models. We utilize LIF neuron in a full-precision manner to keep the information from the input patches. Moreover, we propose the horizontal LIF and vertical LIF to inhert the knowledge in different directions and the group LIF to extract better local features. Experiments on classification, segmenation and detection show that the proposed SNN-MLP models can achieve the state-of-the-art performance among existing MLPs. Especially, the proposed model achieves 81.9\%, 83.3\% and 83.5\% top-1 accuracy on ImageNet dataset with only 4.4G, 8.5G and 15.2G FLOPs, respectively. The FLOPs-Accuracy Pareto frontier is shown in Figure~\ref{fig:pareto-frontier}.

\section{Related Works}
\label{sec:related-work}
\subsection{Spiking Neural Networks}
\label{sec:rw-snn}
Spiking neural networks are a kind of brain-spired neural networks. There are multiple spiking neural models, such as leaky integrate and fire (LIF)~\cite{dayan2003theoretical}, Hodgkin-Huxley (H-H)~\cite{hodgkin1952quantitative} and Izhikevich~\cite{izhikevich2003simple}. The LIF model is most commonly used because it is simple and efficient to implement.

Different from CNNs, SNNs are not originally designed with gradient-based supervised learning at first. Traditional ways to train SNNs are spike timing dependent plasticity (STDP)~\cite{masquelier2007unsupervised}, which is a unsupervised learning method. The main disadvantage of STDP is that global information could not be used, which restrict the speed of converge. This leads to the difficulty of its application on large models. Therefore, several gradient-based training methods for SNN are proposed. Wu et al. propose explicitly iterative LIF neuron~\cite{wu2019direct} to make a faster and better training. Zheng et al.~\cite{zheng2020going} propose threshold-dependent batch normalization and further improve the direct training process. The strength of gradient-based training method is that the trained SNN only requires a few time steps, like  $t=6$ or $t=10$, so the latency is acceptable. Unfortunately, although many efforts have been made, there is still a significant accuracy gap between direct-trained SNNs and CNNs.

Another way to obtain an SNN model is to convert the well-trained ANNs/CNNs into SNNs. This conversion can almost maintain the accuracy of original ANNs/CNNs. Non-spiking ANNs/CNNs are trained normally at first and then converted to spiking neurons~\cite{rueckauer2017conversion, sengupta2019going} by counting the fire rate. Recently, there are some work to combine conversion and training process, such as progressive conversion~\cite{severa2019training} and conversion as an initialization~\cite{rathi2020enabling}. However, to compensate for the loss of accuracy in converting from full precision to binary output, this conversion process always requires large time steps, so it is difficult to achieve satisfying latency. Besides, the conversion algorithm has poor performance on extra deep neural networks. To alleviate these problems, Li et al.~\cite{li2021free} propose a calibration method to improve the accuracy of converted SNN under fewer steps, like $T=128$ or $T=256$. Even though, SNN suffers from 7\% accuracy loss on deep neural network MobileNet with $T=128$.

With the development of training or conversion techniques, the application range of SNNs is gradually expanded. Some researchers have been exploring the possibility of applying SNNs to various computer vision tasks, including segmentation~\cite{patel2021spiking, kim2021beyond} and detection~\cite{kim2020spiking}. In the biomedical field, SNNs have received extensive attention on tasks like MRI image segmentation~\cite{ahmadi2021qais} and ECG classification~\cite{yan2021energy}. Meantime, there are many works on hardware platforms of SNN, like TrueNorth~\cite{akopyan2015truenorth} and Loihi~\cite{davies2018loihi}. Though many efforts have been made, the accuracy of SNN is still not the state-of-the-art.

\subsection{Transformers and MLPs}

\begin{figure*}[t]
\centering
\vspace{-10pt}
\includegraphics[width=6in]{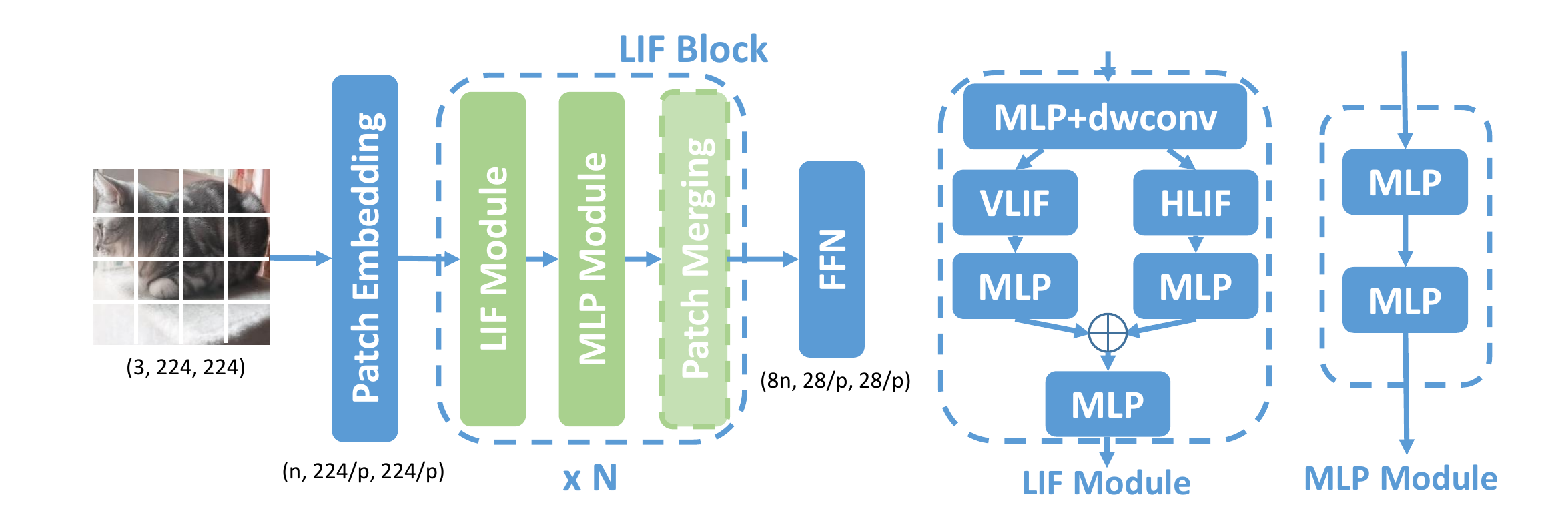}
\vspace{-15pt}
\caption{Framework of our proposed SNN-MLP.}
\vspace{-10pt}
\label{fig:snn-mlp}
\end{figure*}
Transformers~\cite{vaswani2017attention} are widely used in NLP tasks, since they can be highly parallelized. Vision Transformer (ViT)~\cite{dosovitskiy2020image} first introduces Transformer to classification task and applies the Transformer Encoder to extrace features. Transformers are soon used in various computer vision tasks, including detection~\cite{zhu2020deformable} and low-level vision tasks~\cite{chen2021pre}. DeiT~\cite{touvron2021training} proposes to use distillation to improve the training process on the basis of ViT. TNT~\cite{han2021transformer} proposes to model the inner information of patches by embedding Transformers into Transformers. Light-weight transformers are also attracting attention, such as Lite-Transformer~\cite{wu2020lite} and ViT-Lite~\cite{hassani2021escaping}. The latest researches, like CvT~\cite{wu2021cvt} and CMT~\cite{guo2021cmt}, focus on merging CNN and Transformer to absorb the advantage of both architectures.

At the same time, the researchers find that replacing complex multi-head self-attention operations with MLP yields excellent results. In 1980s, MLPs were once all the rage. Now MLPs are different from the older ones, since they need to embed images to patches and then extract features on these patches. MLP-mixer~\cite{tolstikhin2021mlp} takes the lead to claim that MLPs work as well as Transformers. They use a permutation operation to communicate between patches. The general structure of MLPs usually contains two main parts, channel mixing module and token mixing module (permutation, shift and~\etc). The following works are concentrating on improving the token mixing process. ViP~\cite{hou2021vision} permutes on H-C and W-C dimention to extract features. $S^{2}$-MLP~\cite{yu2021s}, cycle-MLP~\cite{chen2021cyclemlp} and AS-MLP~\cite{lian2021mlp} use shift operation to interleave information from different patches. Recently researchers are raising more new ideas to aggregate token information better, like hierarchical re-arrangement~\cite{guo2021hire} and phase-aware representation~\cite{tang2021image}. As a summary, the information communication mechanisms between tokens are important for the performance of MLP models.

\section{Proposed Method}
\subsection{Framework}
\begin{figure*}[t]
\centering
\vspace{-5pt}
\includegraphics[width=6.4in]{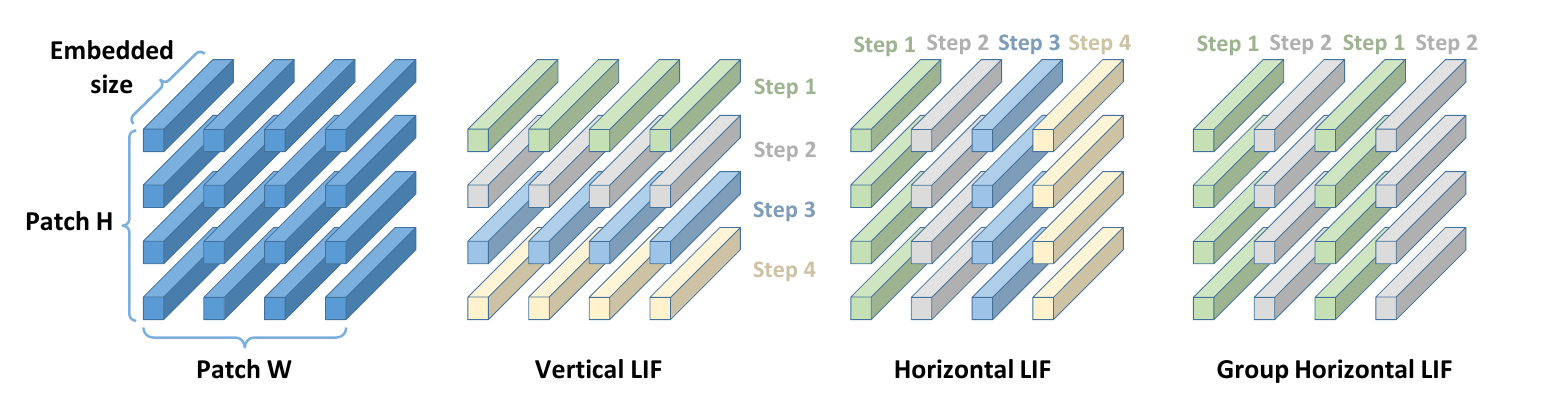}
\vspace{-15pt}
\caption{How we apply LIF neurons to the feature maps.}
\vspace{-15pt}
\label{fig:lif}
\end{figure*}
Here we first present the framework of our SNN-MLP model, which is shown in the left of Figure~\ref{fig:snn-mlp}. The detailed structure of each module will be presented in the following sections. The input image $X$ is divided into patches with size $3\times p \times p$, in which $p$ represents the height and width of patch images, and then a following MLP layer embeds each patch to an $n$ dimension vector. Finally we get a $n\times \frac{H}{p} \times \frac{W}{p}$ feature map. The feature map is fed into our four-stage LIF blocks. One LIF block contains one LIF module and one MLP module. In the LIF block, the LIF module is in charge of the token mixing job while an MLP block is in charge of the channel mixing job. If it is the last block of last three stages, there is also a patch merging module which splits the features of each $2\times2$ neighboring patches to four channels, concatenates them, and then uses a linear layer to reduce the number of channels to a half. Finally, the size of feature maps become $n\times \frac{H}{p\times 2^3}\times \frac{W}{p\times 2^3}$. The classifier then generates the probability vector from the final feature map.

The variants of SNN-MLP has different embedded dimensions and number of blocks. The embedded dimensions $n$ of tiny, small and base model are 96, 96, 128, respectively. The number of blocks of stage 1, 2 and 4 is 2 for all variants, while the number of blocks of stage 3 is 6, 18, 18 for different sizes.

\subsection{Full-precision LIF}
\label{sec:fplif}
In this section, we give a brief introduction of the traditional LIF neuron and our modification on that. The behavior of classical LIF model can be modeled as follows
\begin{align}
o=0,&\tau\frac{du}{dt}=-u+I,u<V_{th}\\
o=1,&u=u_{reset},u\ge V_{th}
\end{align}
where $u$ is the membrane potential, $I$ is the input from the upper layers, $\tau$ is a time co-efficient, $o$ is the output and $V_{th}$ is the fire threshold of this neuron. While a spiking is fired, the membrane potential $u$ is reset to $u_{reset}$.

Many efforts have been made to apply LIF neurons to deep neural networks, the most successful trial is iterative LIF~\cite{wu2019direct}.
\begin{align}
y_{t+1}^{n} &= \sum W^Tx,\\
u_{t+1}^{n} &= \tau u_{t}^{n}(1 - o_t^{n}) + y_{t+1}^{n},\\
o_{t+1}^n &= u_{t+1}^n > V_{th}.
\end{align}

The subscript $t$ represents the time step and superscript $n$ represents the layer index. The element $W$, $x$ and $y$ represents the weight, input and output, respectively. In this paper, we want to adopt LIF machenism as a kind of token mixing method. Different from the accumulation on the time domain in the traditional LIF neurons, we accumulate and fire on the spatial domain instead. The $t$ in the formula represents the index of patches instead of time steps in our design. Since the input feature is full-precision, we prefer a full-precision output to keep the information in the patches. To meet our needs, we propose the following full-precision LIF function:
\begin{align}
o=0,&\tau\frac{du}{dt}=-u+I,u<V_{th}\\
o=u,&u=u_{reset},u\ge V_{th}
\end{align}

We replace the output $1$ with $u$, so the full-precision information is reserved. Applying our full-precision LIF model to the iterative LIF and then we get:
\begin{align}
y_{t+1}^{n} &= \sum W^Tx,\\
u_{t+1}^{n} &= \tau u_{t}^{n}(1 - o_t^{n}) + y_{t+1}^{n},\\
o_{t+1}^n &= u_{t+1}^n > V_{th},\\
r_{t+1}^n &= max(u_{t+1}^n, V_{th}).
\end{align}

Noted that, $r_{t+1}^{n}$ is the final full-precision output of $t+1$ step and $o_{t+1}^n$ is only a temporary variable to record the output state of $t+1$ step. The coefficient $\tau$ and $V_{th}$ are learnable. We discuss the initialization of $\tau$ and $V_{th}$ in Section~\ref{sec:abl}. Then we get a full-precision iterative LIF neuron which can communicate between patches since different patches are regarded as different time steps.

With this explicitly iterative LIF neuron, the backpropagation process can be finished by chain rule.
\begin{equation}
\begin{aligned}
&\frac{\partial L}{\partial \tau} = \frac{\partial L}{\partial r_{t+1}^n}\frac{\partial r_{t+1}^n}{\partial u_{t+1}^n}\sum_{i=0}^{t-1}(\frac{\partial u_{t+1-i}^n}{\partial \tau}\cdot\prod_{j=t+2-i}^{t+1}\frac{\partial u_{j}^n}{\partial u_{j-1}^n}) \\ 
&=\frac{\partial L}{\partial r_{t+1}^n} o_{t+1}^n \sum_{i=0}^{t}(u_{t-i}^n (1-o_{t-i}^n) \prod_{j=t+2-i}^{t+1}\tau (1-o_{j-1}^n)).
\end{aligned}
\end{equation}
\begin{equation}
\frac{\partial L}{\partial V_{th}} = \frac{\partial L}{\partial r_{t+1}^n} \frac{\partial r_{t+1}^n}{\partial V_{th}} = \frac{\partial L}{\partial r_{t+1}^n}\cdot (1 - o_{t+1}^n).
\end{equation}
\begin{equation}
\begin{aligned}
\frac{\partial L}{\partial y_{t+1}^n} &= \sum_{i=t+1}^{end}\frac{\partial L}{\partial r_i}\frac{\partial r_i}{\partial u_i}\frac{\partial u_{t+1}^n}{\partial y_{t+1}^n}\prod_{j=t+2}^{i}\frac{\partial u_j^n}{\partial u_{j-1}^n}\\
&= \sum_{i=t+1}^{end}\frac{\partial L}{\partial r_i^n}\cdot o_i^n \cdot \prod_{j=t+2}^{i}\tau (1-o_{j-1}^n).
\end{aligned}
\end{equation}

\subsection{LIF Module}
\begin{algorithm}[h]
\vspace{-3pt}
\caption{PyTorch-like code of SNN-MLP}
\label{alg:1}
\begin{lstlisting}[language=Python]
def lif(x, dir=2):
	for step in range(groups):
		if dir == 2:
			u, o, x[:,:,step::groups,:] = 
			  lif(u, o, x[:,:,step::groups,:])
		else:
			u, o, x[:,:,:,step::groups] = 
			  lif(u, o, x[:,:,:,step::groups])
	return x
def lif_module_forward(x):
	x = gelu(norm(mlp(x)))
	x = gelu(norm(dwconv(x)))
	x_v = gelu(mlp(vlif(x)))
	x_h = gelu(mlp(hlif(x)))
	x = mlp(norm(x_v + x_h))
	return x
def mlp_module_forward(x):
	x = dropout(gelu(mlp(x)))
	x = gelu(mlp(x))
	return x
def SNN_MLP_forward(x):
	x = patch_embed_forward(x)
	for i in range(4):
		for j in range(block_num[i]):
			x = lif_module_forward(x)
			x = mlp_module_forward(x)
			if j == 0 and i < 3:
				x = patch_merging_forward(x)
	x = classifier(x)
	return x
\end{lstlisting}
\vspace{-3pt}
\end{algorithm}
\begin{table}[t]
\caption{Comparison with state-of-the-art Transformer-based and MLP-based models. Top-1 means the top-1 accuracy on ImageNet1k dataset. The input resolution of all models is $224\times 224$.}
\vspace{-10pt}
\label{table:results}
\centering
\begin{tabular}{c|c|c|c}
\hline\hline
                              & \#Param & \#FLOPs & Top-1 (\%) \\ \hline\hline
gMLP-S~\cite{liu2021pay}                        & 20M     & 4.5G        & 79.4           \\
ResMLP-S24~\cite{touvron2021resmlp}                     & 30M     & 6.0G         & 79.4           \\
DeiT-S~\cite{touvron2021training} & 22M & 4.6G  & 79.8 \\
ViP-Small/14~\cite{hou2021vision}                  & 30M     & -         & 80.5           \\
Swin-T~\cite{liu2021swin} & 29M & 4.5G & 81.3 \\
AS-MLP-T~\cite{lian2021mlp}                      & 28M     & 4.4G        & 81.3           \\
CvT-13~\cite{wu2021cvt} & 20M & 4.5G & 81.6 \\
CycleMLP-B2~\cite{chen2021cyclemlp}                   & 27M     & 3.9G       & 81.6           \\
\textbf{SNN-MLP-T(ours)}      & 28M     & 4.4G       & \textbf{81.9}           \\ \hline
MLP-Mixer-B/16~\cite{tolstikhin2021mlp}                & 59M     & 11.7G       & 76.4           \\
S$^2$-MLP-deep~\cite{yu2021s} & 51M     & 10.5G & 80.7           \\
CvT-21~\cite{wu2021cvt} & 32M & 7.1G & 82.5 \\
ViP-Medium/7~\cite{hou2021vision}                  & 55M     & -          & 82.7           \\
Swin-S~\cite{liu2021swin} & 50M & 8.7G & 83.0 \\
CycleMLP-B4~\cite{chen2021cyclemlp}                   & 52M     & 10.1G       & 83.0           \\
AS-MLP-S~\cite{lian2021mlp}                      & 50M     & 8.5G         & 83.1           \\
\textbf{SNN-MLP-S(ours)}      & 50M     & 8.5G     & \textbf{83.3}           \\ \hline
S$^2$-MLP-wide~\cite{yu2021s} & 71M     & 14.0G       & 80.0           \\
ResMLP-B24~\cite{touvron2021resmlp}                     & 116M     & 23.0G       & 81.0          \\
gMLP-B~\cite{liu2021pay}                          & 73M     & 15.8G    & 81.6           \\
DeiT-B~\cite{touvron2021training} & 86M & 17.5G & 81.8 \\
ViP-Large/7~\cite{hou2021vision}                   & 88M     & -         & 83.2           \\
CycleMLP-B5~\cite{chen2021cyclemlp}                   & 76M     & 12.3G       & 83.2           \\
Swin-B~\cite{liu2021swin} & 88M & 15.4G & 83.5 \\
AS-MLP-B~\cite{lian2021mlp}                      & 88M     & 15.2G    & 83.3           \\
\textbf{SNN-MLP-B(ours)}      & 88M     & 15.2G    & \textbf{83.5}           \\ \hline\hline
\end{tabular}
\vspace{-20pt}
\end{table}

Finally we give an introduction of our LIF module. The structure of our LIF module is shown in Figure~\ref{fig:snn-mlp}. Different from AxialShift block, we add a dwconv after the first MLP layer and replace the shift operations with our LIF neurons. Since the iterative LIF neuron is essentially an activation function, it is necessary to add a dwconv layer to the front of the iterative LIF neuron.

The next problem is to determine the order to communicate between different patches. Following the ideas of previous work~\cite{lian2021mlp}, we communicate information on two directions: vertical and horizontal. The way how Vertical LIF and Horizontal LIF neuron work is shown in Figure~\ref{fig:lif}. The dimension of feature maps are $(N, C, H, W)$, while $C$ refers to the embedded dimension of each patch and $H$/$W$ refer to the height/width of patches. The middle and right subfigures show our Vertical LIF and Horizontal LIF process. Taking VLIF as an example, the first row is referred as $y^n$ in formula (6). Then $u_{1}^n$ equals to $y^n$ and we get $o_1^{n}$ by compare $u_{1}^n$ with $V_{th}$. The output $r_{1}^n$ is only related to $u_1^{n}$ and $V_{th}$ and the tensor $o_1^{n}$ is just used to determine whether $u_1^{n}$ is accumulated to $u_2^{n}$. For the second step, we conduct the same operation on the second row. The only difference is that $u_2^n$ has accumulated values from part of $u_1^n$ which is less than $V_{th}$. Repeat these steps and we get the final results. The processing of HLIF is the same, except that the accumulated vectors become column vectors. A PyTorch-style code of our SNN-MLP model is shown in Algorithm~\ref{alg:1}. In practice, we apply GroupNorm following~\cite{lian2021mlp}.
\vspace{10pt}

\subsection{Group LIF}
For traditional SNNs, excessive time steps may lead to the bad performance~\cite{he2020comparing}. In our practice, we also find that the global LIF performs bad sometimes, and we think this may be due to the introduction of long-distance information with weak correlations. Moreover, excessive time steps may affect the efficiency of parallel computing. To avoid this problem, we divide the feature maps into several groups and apply LIF neurons to each group. With group LIF neurons, closer patterns can communicate with each others. A simple example of $group=2$ horizontal LIF is shown in the right of Figure~\ref{fig:lif}. The $2n$th column, $n\in \mathbb{N}$, is sent to the LIF neuron first. The values larger than $V_{th}$ are to be retained, and the others are set to $V_{th}$ while their values are accumulated to the corresponding element in the next column.
\begin{table*}[ht]
\vspace{-12pt}
\caption{Results on COCO datasets}
\vspace{-10pt}
\label{table:det-res}
\centering
\begin{tabular}{c|c|cccccccc}
\hline\hline
Method                                                                           & Backbone      & AP$^b$ & AP$_{50}^b$ & AP$_{75}^b$ & AP$^m$ & AP$^m_{50}$ & AP$^m_{75}$ & \#Param & \#FLOPs \\ \hline\hline
\multirow{10}{*}{\begin{tabular}[c]{@{}c@{}}Mask\\ R-CNN~\cite{he2017mask}\end{tabular}}            & ResNet-50~\cite{he2016deep}     & 41.0   & 61.7   & 44.9   & 37.1   & 58.4   & 40.1   & 44M    & 260G  \\                                                                                 & PVT-Small~\cite{liu2021swin}        & 43.0   & 65.3   & 46.9   & 39.9   & 62.5   & 42.8   & 44M    & 245G  \\
                                                                                 & Swin-T~\cite{liu2021swin}        & 46.0   & 68.2   & 50.2   & 41.6   & 65.1   & 44.8   & 48M    & 264G  \\
                                                                                 & AS-MLP-T~\cite{lian2021mlp}      & 46.0   & 67.5   & 50.7   & 41.5   & 64.6   & 44.5   & 48M    & 260G  \\
                                                                                 & SNN-MLP-T     & 46.0   & 67.9   & 50.9   & 41.6   & 64.9   & 44.7   &  48M      &  261G     \\ \cline{2-10} 
                                                                                 & ResNet-101~\cite{he2016deep}    & 42.8   & 63.2   & 47.1   & 38.5   & 60.1   & 41.3   & 63M    & 336G  \\                                                                                 & PVT-Medium~\cite{liu2021swin}        & 44.2   & 66.0   & 48.2   & 40.5   & 63.1   & 43.5   & 64M    & 305G  \\
                                                                                 & Swin-S~\cite{liu2021swin}        & 48.5   & 70.2   & 53.5   & 43.3   & 67.3   & 46.6   & 69M    & 354G  \\
                                                                                 & AS-MLP-S~\cite{lian2021mlp}      & 47.8   & 68.9   & 52.5   & 42.9   & 66.4   & 46.3   & 69M    & 346G  \\
                                                                                 & SNN-MLP-S     & 48.0   & 69.1   &52.6    & 42.8   & 66.2   & 46.3   &    69M    &   346G    \\ \hline\hline
\multirow{10}{*}{\begin{tabular}[c]{@{}c@{}}Cascade\\ Mask\\ R-CNN~\cite{chen2019hybrid}\end{tabular}} & DeiT-S~\cite{touvron2021training}         & 48.0   & 67.2   & 51.7   & 41.4   & 64.2   & 44.3   & 80M       & 889G      \\
                                                                                 & ResNet-50~\cite{he2016deep}     & 46.3   & 64.3   & 50.5   & 40.1   & 61.7   & 43.4   & 82M       & 739G      \\
                                                                                 & Swin-T~\cite{liu2021swin}        & 50.5   & 69.3   & 54.9   & 43.7   & 66.6   & 47.1   &   86M     & 745G      \\
                                                                                 & AS-MLP-T~\cite{lian2021mlp}      & 50.1   & 68.8   & 54.3   & 43.5   & 66.3   & 46.9   & 86M       & 739G      \\
                                                                                 & SNN-MLP-T     & 50.3   & 68.9   & 54.6   & 43.6   & 66.5   & 47.1   &   86M     &  739G     \\ \cline{2-10} 
                                                                                 & ResNeXt101-32~\cite{xie2017aggregated} & 48.1   & 66.5   & 52.4   & 41.6   & 63.9   & 45.2   & 101M       & 819G      \\
                                                                                 & Swin-S~\cite{liu2021swin}        & 51.8   & 70.4   & 56.3   & 44.7   & 67.9   & 48.5   & 107M       & 838G      \\
                                                                                 & AS-MLP-S~\cite{lian2021mlp}      & 51.1   & 69.8   & 55.6   & 44.2   & 67.3   & 48.1   & 107M       & 824G      \\
                                                                                 & SNN-MLP-S     & 51.4   & 70.0   &   55.6 & 44.4   & 67.3   & 48.3   &    107M    &   825G    \\ \hline\hline
\end{tabular}
\vspace{-15pt}
\end{table*}
\section{Experiments}
\label{sec:exp}
We conduct experiments on classification, detection and segmentation tasks to show the effectiveness of our SNN-MLP models. All codes are implemented with Python-3.6, PyTorch-1.7~\cite{paszke2017automatic} and MindSpore-1.5~\cite{mindspore}.
\subsection{Classification}
We report our experimental results on ImageNet-1k dataset. ImageNet-1k contains about 1.28M training images and 50K validation images, which are divided into 1000 classes. These images are all RGB images with various shapes. We follow the general input transformations to resize the shorter side to 256 and then crop the whole image to $224\times 224$.

Our training strategy follows Swin Transformer~\cite{liu2021swin}. We use AdamW~\cite{loshchilov2017decoupled} optimizer to train our model for 300 epochs. The initial learning rate is 0.001 with cosine decay. The first 20 epochs are used to warm up the training process. The models are trained on 8 GPUs with batch size 1024. The weight decay is set as 0.05. We also apply label smooth and drop path techniques during the training process.

We evaluate our models (SNN-MLP-T, SNN-MLP-S and SNN-MLP-B) on ImageNet1k. The results are shown in Table~\ref{table:results}. We divide models into three groups according to their parameters and FLOPs. For the tiny models, the number of paramters and FLOPS is below 30M and 6.0G, respectively. Our SNN-MLP-T model achieves 81.9\% top-1 accuracy, which beats other models including gMLP-S, ResMLP-S24, ViP-Small/14, AS-MLP-T and CycleMLP-B2. For the small models, the number of parameters and FLOPs is below 60M and 12.0G. Our SNN-MLP-S model also beats other models with 83.3\% top-1 accuracy. For the large models, our SNN-MLP-B model achieves 83.5\% top-1 accuracy. Compared with AS-MLP models, the number of FLOPs and parameters is all the same since our framework is almost the same as AS-MLP models and the LIF modules introduce little extra computing, but the accuracy is improved by 0.6\%, 0.2\% and 0.2\% for tiny, small and base model, respectively.

\subsection{Detection}
We evaluate our models on COCO dataset, which contains 118K training data and 5K validation data. We implement our models with mmdet-v2.11~\cite{mmdetection} framework and we evaluate our backbone on two famous detection methods: Mask R-CNN~\cite{he2017mask} and Cascade R-CNN~\cite{chen2019hybrid}. To compare fairly, we train our models with the same training strategies as AS-MLP models. We use AdamW optimizer with 0.0001 initial learning rate, and the batch size is set as $2\times 8$ GPUs. The weight decay is 0.05. We also apply the multi-scale training strategy to scale the shorter side between 480 and 800 and the longer side at most 1333. For evaluation, we use single scale $(800, 1333)$ without flipping. Pre-trained models on ImageNet-1k are used to initialize the backbones and then models are trained 36 epochs on COCO training sets (3x schedule).

The results are shown in Table~\ref{table:det-res}. Our SNN-MLP models have almost the same \#FLOPs and \#Param as AS-MLP models. Our SNN-MLP-T and SNN-MLP-S achieve 46.0 AP$^b$ / 41.6 AP$^m$ and 47.9 AP$^b$ / 42.7 AP$^m$ with Mask R-CNN method, and achieve 50.3 AP$^b$ / 43.6 AP$^m$ and 51.4 AP$^b$ / 44.4 AP$^m$ with Cascade Mask R-CNN method, respectively. We can find that our SNN-MLP backbones achieve better results than AS-MLP models, and are also comparable with state-of-the-art backbone Swin-Transformer.
\begin{table*}[ht]
\vspace{-12pt}
\caption{Results on ADE20K datasets}
\vspace{-8pt}
\label{table:seg-res}
\centering
\begin{tabular}{cc|c|c|cc}
\hline\hline
\multirow{2}{*}{Method}                   & \multirow{2}{*}{Backbone}                    & \multicolumn{2}{c|}{Val} & \multirow{2}{*}{\#Param} & \multirow{2}{*}{\#FLOPs} \\ \cline{3-4}
                         &                             & ms mIoU & ss mIoU & & \\ \hline\hline
DANet~\cite{fu2019dual}                    & \multirow{6}{*}{ResNet-101~\cite{he2016deep}} & 45.2 &  -   & 69M     & 1119G   \\
Dlab.V3+~\cite{chen2018encoder}                 &                             & 44.1 &  -   & 63M     & 1021G   \\
ACNet~\cite{fu2019adaptive}                    &                             & 45.9  &  -  & -       & -       \\
DNL~\cite{yin2020disentangled}                      &                             & 46.0  & -   & 69M     & 1249G   \\
OCRNet~\cite{yuan2020object}                   &                             & 45.3   & -  & 56M     & 923G    \\
UperNet~\cite{xiao2018unified}                  &                             & 44.9   & -  & 86M     & 1029G   \\ \hline
OCRNet~\cite{yuan2020object}                   & HRNet-w48~\cite{wang2020deep}                   & 45.7   & -  & 71M     & 664G    \\
DLab.v3+~\cite{chen2018encoder}                 & ResNeSt-101~\cite{zhang2020resnest}                 & 46.9  &  -  & 66M     & 1051G   \\
DLab.v3+~\cite{chen2018encoder}                 & ResNeSt-200~\cite{zhang2020resnest}                 & 48.4   & -  & 88M     & 1381G   \\
SETR~\cite{zheng2021rethinking}                     & T-Large                     & 50.3   & -  & 308M    & -       \\ \hline
\multirow{7}{*}{UperNet~\cite{xiao2018unified}} & DeiT-S~\cite{touvron2021training}                      & 44.0   & -  & 52M     & 1099G   \\
                         & Swin-T~\cite{liu2021swin}                      & 45.8   &  44.5 & 60M     & 945G    \\
                         & Swin-S~\cite{liu2021swin}                      & 49.5    & 47.6 & 81M     & 1038G   \\
                         & Swin-B~\cite{liu2021swin}                      & 49.7   &  48.1 & 121M    & 1188G   \\
                         & AS-MLP-T~\cite{lian2021mlp}                    & 46.5    & - & 60M     & 937G    \\
                         & AS-MLP-S~\cite{lian2021mlp}                    & 49.2   & -  & 81M     & 1024G   \\
                         & AS-MLP-B~\cite{lian2021mlp}                    & 49.5   & -  & 121M    & 1166G   \\ \hline
\multirow{3}{*}{UperNet~\cite{xiao2018unified}} & SNN-MLP-T                   &  46.5   &  45.6    &    60M     &  937G       \\
                         & SNN-MLP-S                   &  49.0   &  48.1    &  81M       &  1025G       \\
                         & SNN-MLP-B                   &  49.4   &  48.4    &   121M      &    1167G     \\ \hline\hline
\end{tabular}
\vspace{-5pt}
\end{table*}
 
\subsection{Sementic Segmentation}
We conduct experiments on the ADE20K, which is a widely-used semantic segmentation dataset. ADE20K contains 20K training images and 2K evaluation images. We choose UperNet method to compare with AS-MLP and implement with mmseg-v0.11~\cite{mmseg2020} framework. Same as AS-MLP, we use AdamW optimizer, $6e^{-5}$ initial learning rate $2\times 8$ GPUs and 0.01 weight decay. For data aumentation, we apply random resize with ratio range $(0.5, 2.0)$, random flip with probability 0.5, random crop with max ratio 0.75 and photo metric distortion. The input images are finally cropped as the $512\times 512$ resolution. While evaluating, we apply multi-scale augmentation and the ratios are set as $(0.5,0.75,1.0,1.25,1.5,1.75)$. we train the models for 160K iterations and ImageNet-1k pre-trained models are used to initialize the backbones.

The results are shown in Table~\ref{table:seg-res}. Our SNN-MLP-T, SNN-MLP-S and SNN-MLP-B models achieve 46.5, 49.0 and 49.4 multi-scale mIoU, which is comparable with AS-MLP models. Besides, our models achieve 45.6, 48.1 and 48.4 single-scale mIoU, which is obviously better than the Swin Transformer backbone.

\begin{table*}[h]
\caption{Comparison with traditional SNN models}
\vspace{-8pt}
\label{table:tra-snn}
\centering
\begin{tabular}{ccccccc}
\hline
Model              & Method             & Activation & Timestep & \#param & FLOPs & Accuracy        \\ \hline
ResNet-34~\cite{he2016deep}          & ANN-SNN conversion~\cite{hu2018spiking} & binary     & 768      & 22M     & 3.7G  & 71.6\%          \\
ResNet-34~\cite{he2016deep}          & Hybrid Training~\cite{rathi2020enabling}    & binary     & 250      & 22M     & 3.7G  & 61.5\%          \\
ResNet-34~\cite{he2016deep}          & STBP-tdBN~\cite{zheng2020going}          & binary     & 6        & 22M     & 3.7G  & 63.7\%          \\
ResNet-34~\cite{he2016deep}          & Calibration~\cite{li2021free}        & binary     & 256      & 22M     & 3.7G  & 74.6\%          \\
RegNetX-4GF~\cite{radosavovic2020designing}        & Calibration~\cite{li2021free}        & binary     & 256      & 21M     & 4.0G  & 77.5\%          \\
\textbf{SNN-MLP-T} & -                  & fp32       & -        & 28M     & 4.4G  & \textbf{81.9\%} \\ \hline
\end{tabular}
\vspace{-10pt}
\end{table*}
\subsection{Ablation Study}
\label{sec:abl}
In this section, we explore the best settings of hyper-parameters, including group of LIF neurons, LIF parameter $V_{th}$ and $\tau$. The effectiveness of our LIF module is also evaluated. All experiments are based on SNN-MLP-T model.

\begin{figure*}[t]
\centering
\includegraphics[width=5.4in]{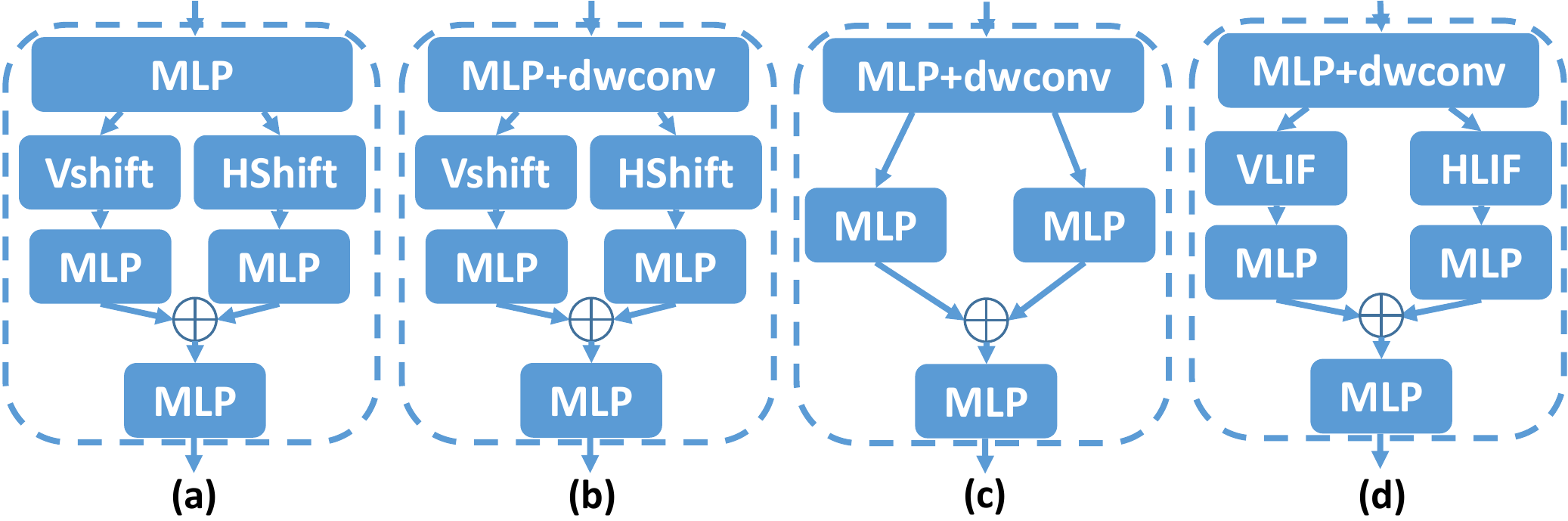}
\vspace{-12pt}
\caption{Modules compared in the ablation study.}
\vspace{-14pt}
\label{fig:ablation-lif}
\end{figure*}
\begin{table}[h]
\vspace{-6pt}
\caption{Ablation study of LIF neuron}
\vspace{-10pt}
\centering
\label{table:abl-lif}
\begin{tabular}{c|ccc|c}
\hline
Index & Dwconv & AxialShift & FP LIF & Accuracy \\ \hline
(a)       &           &    $\surd$          &  & 81.3\%~\cite{lian2021mlp}                \\
(b) &  $\surd$      &     $\surd$         &            & 81.66\%               \\
(c)&   $\surd$     &             &            & 81.56\%               \\
(d)&   $\surd$     &             &     $\surd$        & 81.87\%               \\ \hline
\end{tabular}
\vspace{-11pt}
\end{table}

We train and evaluate four types of models, shown in Figure~\ref{fig:ablation-lif}, and the results are shown in Table~\ref{table:abl-lif}. The original AS-MLP-T model achieves only 81.3\% Top-1 accuracy, while the dwconv+shift module (Figure~\ref{fig:ablation-lif}(b)) achieves 81.66\%. We can find that our dwconv and LIF neurons together achieve the best accuracy 81.87\%, which validates the effectiveness of both of them.
\begin{table}[h]
\vspace{-5pt}
\caption{Ablation study of $\tau$ and $V_{th}$}
\vspace{-10pt}
\centering
\label{table:abl-tau-vth}
\begin{tabular}{c|c|c|c}
\hline
\multirow{2}{*}{Learnable} & \multicolumn{2}{c|}{Init} & \multirow{2}{*}{Top-1 Accuracy} \\ \cline{2-3}
                           & $\tau$         & $V_{th}$        &                                 \\ \hline
$\times$                          &      0.25       &    0        &             81.49                    \\
$\times$                          &      0.25       &   0.25         &           80.98                      \\
$\surd$                          &        0.25     &       0     &                   81.68              \\
$\surd$                              &   0.25          &      0.25      &          \textbf{81.87}                       \\
$\surd$                          &        0.5     &       0.5     &                   81.52              \\ \hline
\end{tabular}
\vspace{-18pt}
\end{table}

The comparison of different $\tau$ and $V_{th}$ settings is shown in Table~\ref{table:abl-tau-vth}. We can see that learnable $\tau$ and $V_{th}$ are significantly better than unlearnable ones. For learnable values, reasonable initial values also make differences on the final results. From the results of our experiments, the initial value 0.25 for both $\tau$ and $V_{th}$ is better than others. We apply 0.25 / 0.25 to all our classification, detection and segmentation experiments.
\begin{table}[h]
\vspace{-5pt}
\caption{Comparison of different LIF groups}
\vspace{-9pt}
\centering
\label{table:lif_groups}
\begin{tabular}{c|cccc}
\hline
LIF groups    & 2     & 4     & 7     & Inf   \\ \hline
Accuracy (\%) & 81.60 & 81.87 & 81.68 & 81.53 \\ \hline
\end{tabular}
\vspace{-14pt}
\end{table}

We also explore the optimal hyper-parameter $g$, which represents the group number of LIF neuron. We evaluate the results under several different $g$ and the results are shown in Table~\ref{table:lif_groups}. The Inf in the table means that we apply global LIF instead of group LIF. We can find that the performance of global LIF significantly worse than group LIF, and the highest accuracy shows when $g=4$. So we adapt $g=4$ in all other experiments.

\subsection{Comparison with SNNs}
Here we also make a brief comparison between our SNN-MLP models and traditional SNN models, shown in Table~\ref{table:tra-snn}. As we mentioned in Section~\ref{sec:rw-snn}, conversion-based methods, including ANN-SNN conversion~\cite{hu2018spiking} and calibration~\cite{li2021free}, may achieve acceptable accuracy while large time steps (over 250 steps) are required. This leads to the unacceptable latency. Another way is to train the SNN models directly, like STBP series, which requires less time steps but get worse performance on accuracy. Compared with traditional SNN methods, at the cost of using full-precision activations instead of binary activations, our SNN-MLP models achieve much better accuracy without the requirement of time steps.

\subsection{Discussion of Various LIF Modules}
\begin{table}[h]
\vspace{-10pt}
\caption{Comparison of various LIF modules}
\vspace{-5pt}
\label{table:var-lif}
\centering
\begin{tabular}{c|ccc}
\hline
Method   & Add LIF & Sub GELU & \textbf{Sub AS} \\ \hline
Accuracy & 81.38\%  & 81.21\%   & \textbf{81.87\%} \\
\hline
\end{tabular}
\vspace{-10pt}
\end{table}

\begin{figure*}[b]
\centering
\vspace{-12pt}
\includegraphics[width=5.8in]{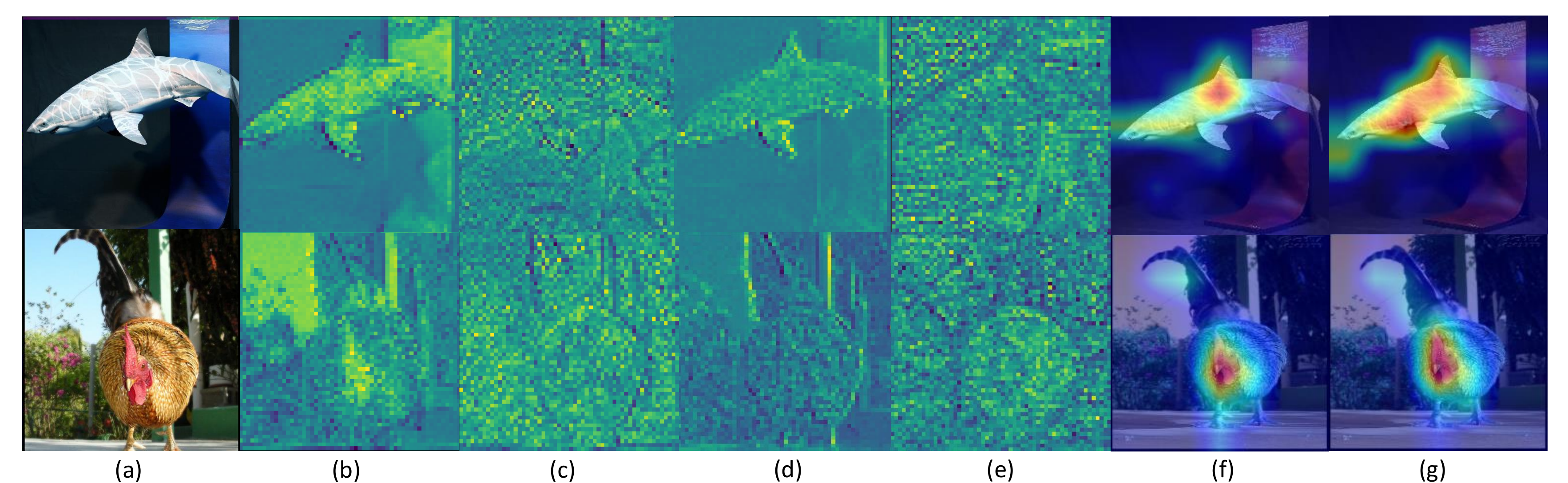}
\vspace{-10pt}
\caption{Visualization results. (a) Original image. (b)(c) Feature maps of the 1st/2nd blocks of AS-MLP-T model. (d)(e) Feature maps of the 1st/2nd blocks of SNN-MLP-T model. (f)(g) GradCAM results of AS-MLP-T and SNN-MLP-T.}
\vspace{-12pt}
\label{fig:vis}
\end{figure*}
We also make attempts to incorporate LIF neurons with MLP in different ways. We try to directly add LIF neurons to the end of each stage of AS-MLP models or subtitute GELU activations in MLP modules. Both choices are not as good as our final models, which actually replace the axial shift modules. A comparison experiment based on AS-MLP-T model is shown in Table~\ref{table:var-lif}. We can find that add LIF neurons and substitute GELUs only achieve 81.38\% and 81.21\%, respectively, which is much lower than our final choice.

\subsection{Visualization}
We provide some visualization results to help understand our design. The leaky-fire process actually removes some noise and unimportant information, and the integrate process makes some compensations to avoid the complete loss of information. From Figure~\ref{fig:vis}, we can find that the LIF neurons extract better texture features compared with AxialShift modules, and the AxialShift modules just make it more like the original images.

\section{Conclusions}
In this paper, we incorporate the machanism of LIF neurons into the MLP models to further improve their accuracies. We propose a full-precision LIF operation to communication between patches, in which we replace the time step with the spatial patches. Besides, we propose the group LIF to extract better local features. With these methods, we design the LIF module which contains horizontal LIF and vertical LIF to deal with features in different directions. We evaluate our methods on different computer vision tasks, including classification, detection and sementic segmentation. On ImageNet-1k dataset, our SNN-MLP models achieve 81.9\%, 83.3\% and 83.5\% top-1 accuracy for different scales, and all of them are higher than the base AS-MLP models. For detection and sementic segmentation tasks, we evaluate on the COCO and ADE20k dataset respectively, and all achieve comparable results with state-of-the-art backbones, including AS-MLP and Swin-Transformer. Finally, we conduct several ablation studies to show the effectiveness of our methods.

In the future, we would continue to explore and improve the utility of our LIF neurons in various vision tasks including detection and sementic segmentations. And we would also try to incorporate LIF neurons with more MLP and Transformer backbones.

{\small
\bibliographystyle{ieee_fullname}
\bibliography{PaperForReview-camera}
}

\end{document}